\documentclass[conference,a4paper]{IEEEtran}
\IEEEoverridecommandlockouts
\usepackage{cite}
 \usepackage{color}

\usepackage{float}
\usepackage{derivative}
\usepackage{amsmath,amssymb,amsfonts}
\usepackage[english]{babel}
\usepackage{amsthm}
\theoremstyle{definition}

\usepackage{graphicx}
\usepackage{caption}
\usepackage{subcaption}
\usepackage{multirow}
\usepackage[left=1.35cm,right=1.35cm,top=1.9cm,bottom=4.1cm]{geometry}
\usepackage{textcomp}
\usepackage{xcolor}
\usepackage{bbm}
\usepackage{algorithm}
\usepackage{algpseudocode}

\def\BibTeX{{\rm B\kern-.05em{\sc i\kern-.025em b}\kern-.08em
    T\kern-.1667em\lower.7ex\hbox{E}\kern-.125emX}}
\usepackage{tikz}
\begin{document}
\def\blue{\textcolor{blue}}
\title{Graph Neural Networks for Resource Allocation in Multi-Channel Wireless Networks\\

}
\newcommand\blfootnote[1]{%
  \begingroup
  \renewcommand\thefootnote{}\footnote{#1}%
  \addtocounter{footnote}{-1}%
  \endgroup
}
\author{\IEEEauthorblockN{Lili Chen$^{\ast}$, Changyang She$^{\dagger}$, Jingge Zhu$^{\ast}$ and Jamie Evans$^{\ast}$}

\IEEEauthorblockA{$^{\ast}$Department of Electrical and Electronic Engineering, University of Melbourne, Australia \\
$^{\dagger}$School of Information Science and Technology, Harbin Institute of Technology (Shenzhen), Shenzhen, China.\\
Email:\{lili.chen1, jingge.zhu, jse\}@unimelb.edu.au, shechangyang@gmail.com\\
}

}
\maketitle
\begin{abstract}
As the number of mobile devices continues to grow, interference has become a major bottleneck in improving data rates in wireless networks. Efficient joint channel and power allocation (JCPA) is crucial for managing interference. In this paper, we first propose an enhanced WMMSE (eWMMSE) algorithm to solve the JCPA problem in multi-channel wireless networks. To reduce the computational complexity of iterative optimization, we further introduce JCPGNN-M, a graph neural network-based solution that enables simultaneous multi-channel allocation for each user.  We reformulate the problem as a Lagrangian function, which allows us to enforce the total power constraints systematically. Our solution involves combining this Lagrangian framework with GNNs and iteratively updating the Lagrange multipliers and resource allocation scheme. Unlike existing GNN-based methods that limit each user to a single channel, JCPGNN-M supports efficient spectrum reuse and scales well in dense network scenarios. Simulation results show that JCPGNN-M achieves better data rate compared to eWMMSE. Meanwhile, the inference time of JCPGNN-M is much lower than eWMMS, and it can generalize well to larger networks. 
\end{abstract}

\begin{IEEEkeywords}
Joint Resource Management, Graph Neural Networks, Wireless Communication, Multiple Channels
\end{IEEEkeywords}

\section{Introduction}\label{C5:sec:Introduction}

As the density of mobile devices increases, interference becomes the bottleneck for improving the data rate in future wireless networks. To support more devices, future communication systems will use larger bandwidths and more carriers. Thus, the channel and power allocation policy is pivotal to alleviating the interference in multi-channel wireless networks. 

\subsection{Related Works}

Channel and power allocation have been extensively studied in the literature, with a focus on optimizing resource utilization in diverse wireless communication scenarios. The authors of \cite{shi2011iteratively} converted the Weighted Sum Rate maximization (WSRM) problem into a Weighted Mean-Square Error Minimization (WMMSE) problem, achieving convexity for individual users and simplifying local optimization. 
It is worth noting that WMMSE imposes a restrictive constraint. For example, each transceiver pair can access only one subchannel at a time, leading to suboptimal spectrum efficiency. To improve the achievable rate of the network, multi-channel communications have been explored in \cite{elnourani2018underlay}, where each transceiver pair is permitted to access multiple channels simultaneously. However, \cite{elnourani2018underlay} imposes the restriction that each channel can only be used by one transceiver pair.

To overcome this limitation, the methods in \cite{hajiaghajani2016joint} and \cite{mach2019resource} propose heuristic channel assignment algorithms that allow channels to be shared by multiple transceiver pairs, with each transceiver pair capable of reusing multiple channels. While these approaches enhance spectral efficiency, they have notable drawbacks. Specifically, \cite{hajiaghajani2016joint} relies on fixed transmission power for cellular users, leading to sub-optimal performance in dynamic network conditions. With the method in \cite{mach2019resource}, the sequential channel allocation with interference evaluation incurs high computational complexity, scaling cubically with the number of transceiver pairs \cite{mach2019resource}, which limits their scalability and practical applicability in large networks.

 Graph Neural Networks (GNNs) offer a compelling alternative by leveraging the inherent topology of wireless networks. Recent works \cite{peng2024learning, chen2023graph} applied graph neural networks (GNNs) for power allocation in cell-free massive MIMO and full-duplex networks, aiming to reduce interference and enhance user experience. Extending this direction, \cite{li2024hetero} tackled power allocation in multicarrier-division duplex cell-free massive MIMO systems.
The authors of \cite{chen2021gnn} proposed a GNN to learn the channel allocation given a power allocation scheme. Nonetheless, their methodology imposes a restriction that each channel can be accessed by at most one transceiver pair. Different from \cite{chen2021gnn}, the authors of \cite{chen2024gnn,marwani2024graph} designed a GNN that allows multiple transceiver pairs to access one channel simultaneously.
However, the above schemes \cite{chen2021gnn,chen2024gnn,marwani2024graph} limit each D2D user to access at most one channel. In summary, no existing GNN-based approach provides JCPA for scenarios where users can transmit data over multiple channels simultaneously, which is of high interest, especially in urban areas.

\subsection{Motivation and Contributions}
To address the above issues, this paper delves into the design of a GNN framework for the joint optimization of channel and power allocation in multi-channel wireless networks. The main contributions of this paper are summarized below, 

\begin{itemize}
    \item  
    We develop an enhanced WMMSE (eWMMSE) algorithm, building upon the iterative framework proposed in \cite{shi2011iteratively}, to address the joint channel and power allocation problem in multi-channel wireless networks. 
    \item 
    To reduce computational complexity, we propose a novel Joint Channel and Power allocation GNN algorithm for Multi-channel scenarios (JCPGNN-M). Unlike prior approaches, such as \cite{chen2024gnn,marwani2024graph}, which limit transceiver pairs to accessing at most one channel, JCPGNN-M enables simultaneous multi-channel allocation for each user. This approach aligns with practical systems, facilitating efficient spectrum reuse and scalable operations in dense wireless networks. 
    \item 
    We validate the effectiveness of JCPGNN-M in solving sum-rate maximization problems with unlabeled data, eliminating the need for computationally expensive labeled samples. Extensive simulations demonstrate that JCPGNN-M achieves a higher data rate compared to the eWMMSE algorithm while significantly reducing inference time compared to the convergence time required by iterative optimization algorithms. It also demonstrate excellent generalization capabilities to larger-scale wireless networks.  
\end{itemize}

\section{System Model and Problem Formulation} \label{C5:sec:system}
In this section, we first present the system model and then formulate the joint channel and power allocation problem in a wireless network with multiple channels. 
\subsection{System Model}
\begin{figure}[htbp]
\centerline{\includegraphics[width=8cm]{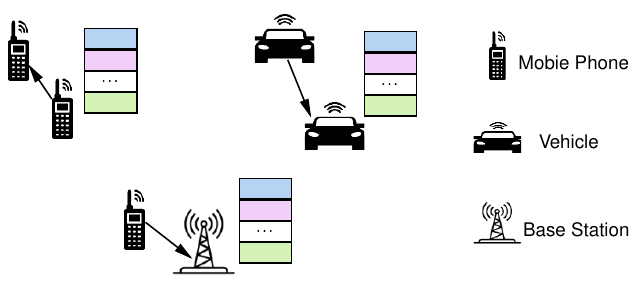}}
\caption{D2D wireless network with multiple channel access.}
\label{C5:fig:systemodel}
\end{figure}

As shown in Figure~\ref{C5:fig:systemodel}, we consider a wireless network with $D$ transceiver pairs denoted by $\mathcal{D} = \{1,2,...,D\}$, where each transmitter or receiver can be a vehicle, base station, or mobile phone.  Mutual interference arises when two transceiver pairs share the same channel. We employ different colors to indicate different channels. Each transceiver pair can access all the channels at the same time. Let $M$ orthogonal channels, each with identical bandwidth, be at the disposal of the system. We represent the index set for these channels as $\mathcal{M} = \{1, 2, ..., M\}$. In this system model, we follow a similar set-up to \cite{nakashima2020deep} in that we refrain from assigning specific values to the bandwidth of the channels but presume the absence of overlap between them. We assume that the bandwidth of each Resource Block (RB) is small enough to exhibit flat-fading channel characteristics. The received signal for the $i$-th receiver at $m$-th channel is given by, 
\begin{equation}
    y_{i}^m = h_{i,i}^m s_{i} c_{i}^m+\sum_{j \neq i}h_{i, j}^m s_{j} c_{j}^m+n_{i}^m,
\end{equation}
where $ h_{i,i}^m \in \mathbb{C} $  and $ h_{i,j}^m \in \mathbb{C} $ represent the channel from transmitter $i$ and $ j $  to receiver $ i $ on the same channel, respectively. The transmitted signal from transmitter $ i $ is denoted by $ s_i \in \mathbb{C} $. The channel allocation is indicated by the binary variable $ c_i^m $, where $ c_i^m = 1 $ if the $ i $-th transceiver pair utilizes the $ m $-th channel for transmission, and $ c_i^m = 0 $ otherwise. The additive Gaussian noise at receiver $ i $ on the $ m $-th channel is modeled as $ n_i^m \sim \mathcal{CN}(0, \sigma_i^2) $. Based on these definitions, the SINR at receiver $ i $ on the $ m $-th channel can be expressed as follows:
\begin{equation}
    \text{SINR}_{i}^{m} = \frac{\left|h_{i,i}^{m}\right|^{2} p_{i}^m c_{i}^{m}}{\sum_{j \neq i}\left|h_{i,j}^{m}\right|^{2} p_{j}^m c_{j}^{m}+\sigma_{i}^{2}},
\end{equation}
where $p_{i}^m$ is the power that the $i$-th transmitter allocates on the $m$-th channel. Denote the channel allocation matrix as $\mathbf{C} = [\mathbf{c}_1,...,\mathbf{c}_D]^{T} \in \mathbb{R}^{D \times M}$, where $\mathbf{c}_i = [c_{i}^1,...,c_{i}^{M}]$ and power allocation matrix as $\mathbf{P} =[\mathbf{p}_1,...,\mathbf{p}_D]^{T} \in \mathbb{R}^{D \times M}$, where $\mathbf{p}_i = [p_{i}^1,...,p_{i}^{M}]$. 
The data rate of the $i$-th transceiver pair over the $m$-th channel is given by,
\begin{equation}
R_{i}^m(\mathbf{C,P})=\log_2 \left(1+\text{SINR}_{i}^{m}\right)\;\text{(bits/s/Hz)}.
\end{equation}
\subsection{Resource Allocation Problem Formulation}
The objective is to maximize the achievable network rate within the system while adhering to the specified sum power constraints. The problem is formulated as follows,
 \begin{subequations}
    \begin{align}
    \underset{\mathbf{C,P}}{\operatorname{maximise}} & \sum_{m=1}^{M}  \sum_{i=1}^{D}  \alpha_{i} R_{i}^m(\mathbf{C,P}), \\
    \text { subject to } & p_i^m\geq0, \quad  \forall i \in \mathcal{D},m \in \mathcal{M},\label{eq:jointchannel_1st} \\
    & c_{i}^m \in \{0,1\}, \quad \forall i \in \mathcal{D}, m \in \mathcal{M},\label{eq:jointchannel_2nd}\\
    &\sum_{m=1}^{M} c_{i}^m p_{i}^m\leq P_{\max }, \quad \forall i \in \mathcal{D},\label{eq:jointchannel_3rd}
    \end{align}
    \label{C5:eq:jointchannel}
\end{subequations}   
where $\alpha_{i}$ is the weight for the $i$-th transceiver pair, $P_{\max }$ denotes the maximum power for transmitter.

\section{Traditional optimization Approach in Resource Allocation Problems}\label{C5:sec:traditional}
\subsection{Optimization Simplification} 
In this section, we extend a traditional optimization approach, WMMSE, to solve the joint channel and power allocation problem. Since $ c_i^m $ is binary, the product $ c_i^m p_i^m $ satisfies the following conditions: (1) If $ c_i^m = 1 $, then $ c_i^m p_i^m = p_i^m $, meaning the channel is assigned to the user with a certain power level. (2) If $ c_i^m = 0 $, then $ c_i^m p_i^m = 0 $, indicating that no power is allocated to that channel. Thus, the sum $ \sum_{m=1}^{M} c_i^m p_i^m $ effectively reduces to the summation of $ p_i^m $ over only those channels where $ c_i^m = 1 $.  In Problem \eqref{C5:eq:jointchannel}, since users are allowed to access multiple channels simultaneously, we observe that a user is considered active on a channel if and only if $ p_i^m \neq 0 $. Therefore, we reformulate the original problem into the following equivalence problem:  
\begin{subequations}\label{eq:jointchannelqos_sim}
\begin{align}
    \underset{\mathbf{P}}{\operatorname{maximize}} & \sum_{m=1}^{M}  \sum_{i=1}^{D}  \alpha_{i} R_{i}^m(\mathbf{P}), \\
        \text { subject to } & p_i^m\geq0, \quad  \forall i \in \mathcal{D},m \in \mathcal{M},\label{eq:jointchannelqos_sim_1st}\\
    &\sum_{m=1}^{M} p_{i}^m\leq P_{\max}, \quad \forall i \in \mathcal{D}, \label{eq:jointchannelqos_sim_2nd} 
\end{align}
\end{subequations}
Due to the interference, the simplified optimization problem remains non-convex and challenging to solve.

\subsection{eWMMSE Approach}
The WMMSE algorithm transforms the SRM problem into a three-dimensional space by leveraging the well-known MMSE-SINR equality \cite{shi2011iteratively}. Originally, WMMSE was designed for optimization problems where the decision variables are beamforming vectors in a single-channel setting. To adapt this approach to our problem, which focuses on power allocation across multiple channels, we modify the algorithm accordingly. Following the approach in \cite{sun2018learning}, we replace $ h_{i,j} $ with $ |h_{i,j}| $ to simplify the implementation. According to Theorem 1 in \cite{shi2011iteratively}, Problem \eqref{eq:jointchannelqos_sim} is equivalent to the following weighted Mean Square Error (MSE) minimization problem:
\begin{subequations}
\begin{align}
    \underset{\mathbf{W,U,V}}{\operatorname{minimise}} & \sum_{m=1}^{M}  \sum_{i=1}^{D}  \alpha_{i}(w_i^m e_i^m - \log(w_i^m)), \\
    \text { subject to } &\sum_{m=1}^{M} (v_{i}^m)^2 \leq P_{\max }, \quad \forall i \in \mathcal{D},
    \end{align}
    \label{C5:eq:jointchannel_mmse}
\end{subequations}
where $v_{i}^m = \sqrt{p_i^m}$ denote the square root of the transmission power of transmitter $i$ at the $m$-th channel, $w_i^m$ is a positive variable and $e_i^m$ is the MSE.
To establish equivalence, we first derive the optimal values of $w_i^m$ and $u_i^m$ by solving the first-order optimality conditions of Problem~\eqref{C5:eq:jointchannel_mmse}:
\begin{subequations}
    \begin{align}
        & (u_{i}^{m})_{\text{opt}}=\frac{|h_{i,i}^{m}| v_{i}^{m}}{\sum_{j=1}^{D}\left|h_{i, j}^{m} v_{j}^{m}\right|^{2}+\left(\sigma_{i}^{m}\right)^{2}}=|h_{i, i}^{m}| v_{i}^{m} \cdot\left(J_{i}^{m}\right)^{-1},\\
        & (w_i^m)_{\text{opt}} = (e_i^m)^{-1},
    \end{align}
    \label{C5:eq:wmmse_u_optimal}
\end{subequations}
where $J_{i}^{m} = \sum_{j=1}^{D}\left|h_{i, j}^{m} v_{j}^{m}\right|^{2}+\left(\sigma_{i}^{m}\right)^{2}$. Substituting the optimal expressions for $w_i^m$ and $u_i^m$ into Problem \eqref{C5:eq:jointchannel_mmse}, we obtain:
\begin{subequations}
    \begin{align}
    \underset{\mathbf{W,U,V}}{\operatorname{maximise}} & \sum_{m=1}^{M}  \sum_{i=1}^{D}  \alpha_{i}\left(\log \left(\frac{1}{1-\left|h_{i,i}^{m} v_{i}^{m}\right|^{2}\left(J_{i}^{m}\right)^{-1}}\right)\right), \\
    \text { subject to } &\sum_{m=1}^{M} (v_{i}^m)^2 \leq p_{\max }, \quad \forall i \in \mathcal{D}.
    \end{align}
    \label{C5:eq:jointchannel_mmse_op}
\end{subequations}
The Problem \eqref{C5:eq:jointchannel_mmse_op} is identical to the original SRM Problem in \eqref{eq:jointchannelqos_sim}. 
Following the steps outlined in Section III of \cite{shi2011iteratively}, the optimal square root of transmission power is given by:
\begin{equation}
v_{i}^{m}=\frac{w_{i}^{m} u_{i}^{m} h_{i,i}^{m}}{\sum_{j=1}^{D} w_{j}^{m}\left|u_{j}^{m} h_{j, i}^{m}\right|^{2}+\lambda_{i}},
\end{equation}
where $\lambda_i \geq 0$ is the multiplier for transceiver pair $i$ that guarantees the total power constraint. 
This result straightforwardly extends the WMMSE-based power allocation from single-channel to multi-channel systems. The primary difference lies in the presence of multiple power variables across different channels, requiring independent updates for each channel.

\section{Graph Neural Networks for Resource Allocation Problems} \label{C5:sec:GNN}
In this section,  we demonstrate the graph representation for our problem and then propose a GNN-based algorithm that allocates resources to maximize the network sum rate.
\subsection{Learning Approach}
Although an iterative optimization algorithm was introduced in the previous section to solve the problem, its computational complexity remains too high for practical implementation. To address this challenge, we develop a learning-based approach to improve efficiency. To determine the optimal solution for Problem \eqref{eq:jointchannelqos_sim}, we begin by formulating its Lagrangian function as follows:
\begin{equation}
\begin{aligned}
\mathcal{L} (\mathbf{P(\mathbf{H})},\lambda) &\triangleq  -\sum_{m=1}^{M}\sum_{i=1}^{D}\alpha_{i} R_{i}^m(\mathbf{P(\mathbf{H})})\\
&+ \sum_{i=1}^{D} \lambda_{i}\left(\sum_{m=1}^{M} p_{i}^{m}-P_{\text {max }}\right),\label{C5:eq:LagrangeP}
\end{aligned}
\end{equation}
where $\lambda_{i}\geq 0$ is the Lagrange multipliers. We define multiplier vectors $\lambda$, where $\lambda = [\lambda_1,...,\lambda_D]$. Here, $\mathbf{H} \in \mathbb{C}^{(M \times D)\times(M \times D)}$ is the CSI matrix of all channels. Since CSI varies across different channels, we define the matrix $\mathbf{H}$ as follows,
\begin{equation}
\mathbf{H} = \begin{pmatrix}
\mathbf{h_1} & 0 & 0 & \cdots & 0 \\
0 & \mathbf{h_2}& 0 & \cdots & 0 \\
0 & 0 & \mathbf{h_3} & \cdots & 0 \\
\vdots & \vdots & \vdots & \ddots & \vdots \\
0 & 0 & 0 & \cdots & \mathbf{h_M}
\end{pmatrix},
\end{equation}
where $\mathbf{h_m} \in \mathbb{C}^{D \times D}$ for  $m=1,2,...,M$, are the CSI matrix of $m$-th channels. 
We then approximate the power allocation policy, $\mathbf{\tilde{P}(\mathbf{H})}$, by a GNN denoted as $\boldsymbol{G}_p(\omega_p;\mathbf{H})$, where $\omega_p$ is the trainable parameters of the GNN. The corresponding data rate with power allocation policy $\mathbf{\tilde{P}(\mathbf{H})}$ is  $\tilde{R}_{i}^m =R_{i}^m(\mathbf{\tilde{P}(\mathbf{H})})$. According to the universal invariant and equivariant graph neural networks theorem \cite{keriven2019universal}, a function defined on graphs can be approximated by a GNN, and the approximation can be uniformly well. Since our problem is non-convex, a locally optimal solution to Problem \eqref{eq:jointchannelqos_sim} can be obtained by solving Problem \eqref{eq:lagarian_gnn}\cite{sun2023unsupervised}. 
 \begin{figure*}[htbp]
\centerline{\includegraphics[width=0.8\textwidth]{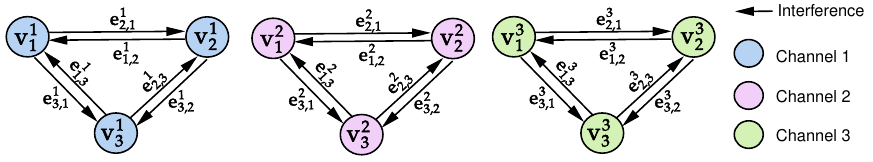}}
\caption{Graph representation of joint resource allocation in multi-channel networks with $D=3$ and $M=3$.}
\label{C5:fig:graphrepre}
\end{figure*}
\begin{equation}
\begin{aligned}
&\underset{\lambda} {\operatorname{max}} \hspace{0.1cm}
\underset{\omega_p} {\operatorname{min}} \hspace{0.2cm}
\hat{\mathcal{L}} (\mathbf{\tilde{P}(\mathbf{H})},\lambda) \\
& \text { s.t. } \lambda_{i} \geqslant 0 \quad i \in D, 
\end{aligned}
\label{eq:lagarian_gnn}
\end{equation}
To solve the problem formulated in \eqref{eq:lagarian_gnn}, a primal-dual method can be employed to iteratively update the primal variables $\omega_p$ and the dual variables $\lambda_i$. 
In the $s$-th iteration, the primal variables $\omega_p$ are updated using the Stochastic Gradient Descent (SGD) method, while the dual variable $\lambda_i$ is updated using the SGA method \cite{sun2023unsupervised}. The update rules are expressed as below, 
\begin{subequations}
\begin{align}
\omega_{p}^{(s+1)} & =\omega_{p}^{(s)}-\phi_{\omega_p} \nabla_{\omega_p} \hat{\mathcal{L}}^{(s)}, \\
\lambda_{i}^{(s+1)} & =\left[\lambda_{i}^{(s)}+\phi_{\lambda_{i}}\left(\sum_{m=1}^{M}\tilde{p}_{i}^{m}-P_{\text {max }}\right)\right]^{+},
\end{align}
\label{eq:multiplier_update}
\end{subequations}
where the operator $[x]^{+} = \text {max}\{0,x\} $ ensures that the dual variables $\lambda_{i} \geqslant 0$. The learning rates for updating the primal variable $\omega_{p}$, the dual variable $\lambda_{i}$ are $\phi_{\omega_p}$ and $\phi_{\lambda_{i}}$ , respectively. The gradient of the Lagrangian $\hat{\mathcal{L}}$ with respect to $\omega_p$ is denoted as $\nabla_{\omega_p} \hat{\mathcal{L}}^{(s)}$.
As demonstrated in \cite{eisen2019learning}, the primal-dual method ensures convergence to a locally optimal solution of the original Problem \eqref{eq:jointchannelqos_sim}.

\subsection{Graph Modeling in Wireless Networks}
To effectively capture the CSI within this framework, we model the $m$-th channel using a subgraph $\mathcal{G}^m$. Since transceiver pairs sharing the same channel interfere with one another, we construct $M$ separate complete graphs as illustrated in Figure~\ref{C5:fig:graphrepre}, ensuring no loss of information.
In the subgraph $\mathcal{G}^m$, the $i$-th vertex $v_i^m$ represents the $i$-th transceiver pair transmitting data over the $m$-th channel. Since interference occurs between transceiver pairs sharing the same channel, we define an interference edge $e_{i,j}^m$ between any two vertices $v_i^m$ and $v_j^m$ in the vertex set $\mathcal{V}$. The neighboring set of vertex $v_i^m$ is denoted as $\mathcal{N}(v_i^m) = \{v_j^m \mid e_{i,j}^m \in \mathcal{E}\}$, which includes all vertices connected to $v_i^m$ by interference edges.

The node features incorporate properties of the transceiver pairs, such as the direct CSI. Specifically, the node feature matrix for vertex $v_i^m$ is defined as $\mathbf{V}_{i}^m = \left[\left|h_{i,i}^m\right|\right]$. The edge features capture the properties of the interference channels. For each interference edge $e_{i,j}^m$, the edge feature vectors are defined as $\mathbf{E}_{i,j}^m = \left[|h_{i,j}^m|, |h_{j,i}^m|\right]$.

\subsection{JCPGNN-M on Graph Neural Networks}

\begin{figure}[htbp]
\centerline{\includegraphics[width=0.3\textwidth]{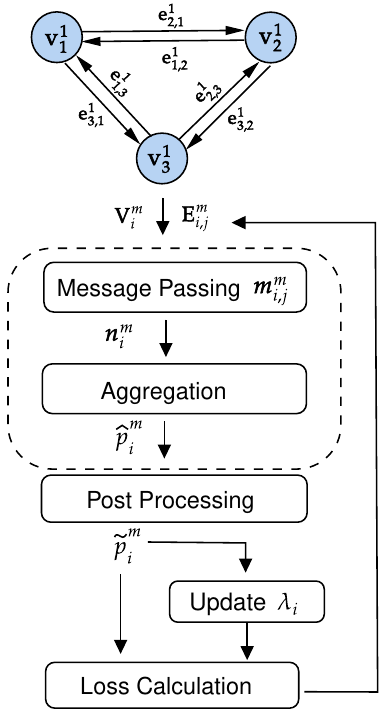}}
\caption{The structure of proposed JCPGNN-M algorithm.}
\label{C5:fig:GNNstructure}
\end{figure}

By modeling a wireless network as a graph $\mathcal{G}$, the goal is to find a function $\boldsymbol{G}_p(\omega_p;\mathbf{H})$ mapping each node $v_i^m$ in the $\mathcal{G}$ to the transmit power allocation $\tilde{p}_i^m(\mathbf{H})$, where $\omega_p$ represents learnable parameters.  
To handle the joint channel and power allocation problem formulated in \eqref{eq:lagarian_gnn}, we propose the JCPGNN-M algorithm, as illustrated in Figure~\ref{C5:fig:GNNstructure}. The JCPGNN-M architecture consists of three main components.
\subsubsection{Message Computation Layer}
Since the users have interference when they share the same channel, the message passing will occur within each subgraph $\mathcal{G}^m$. The update rule for message computation in the $s$-th layer for vertex $v_i^m$ is given by,
\begin{figure*}[t]
\centering
\begin{subfigure}{.32\textwidth}
    \centering
    \includegraphics[width=.95\linewidth]{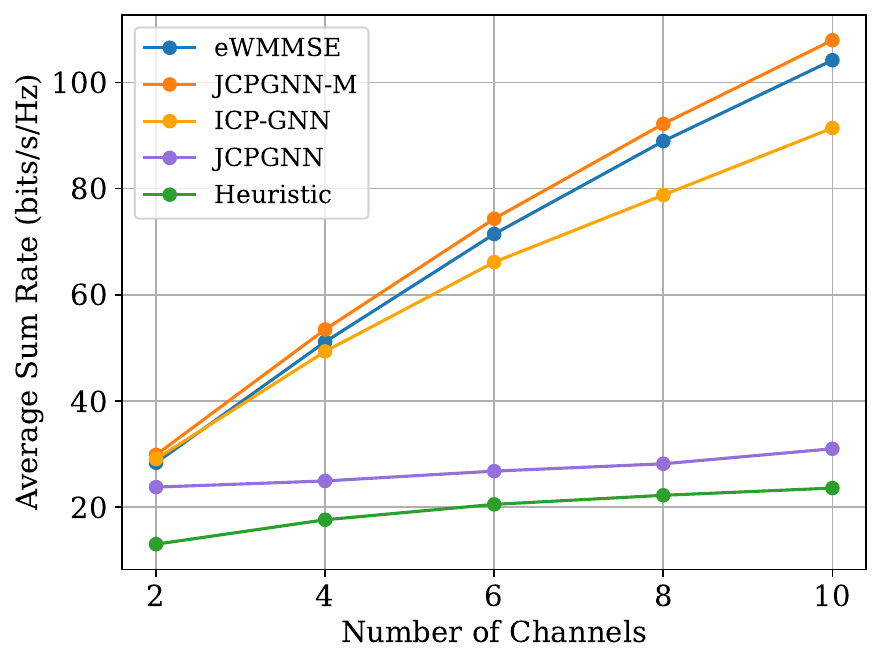}  
    \caption{$D=10$}
\end{subfigure}
\begin{subfigure}{.32\textwidth}
    \centering
    \includegraphics[width=.95\linewidth]{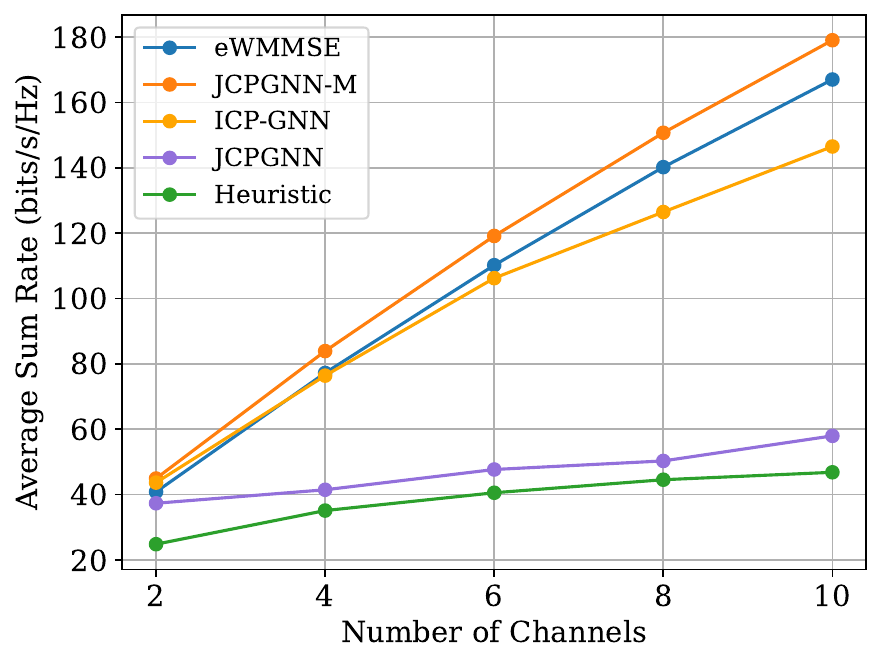}  
    \caption{$D=20$}
\end{subfigure}
\begin{subfigure}{.32\textwidth}
    \centering
    \includegraphics[width=.95\linewidth]{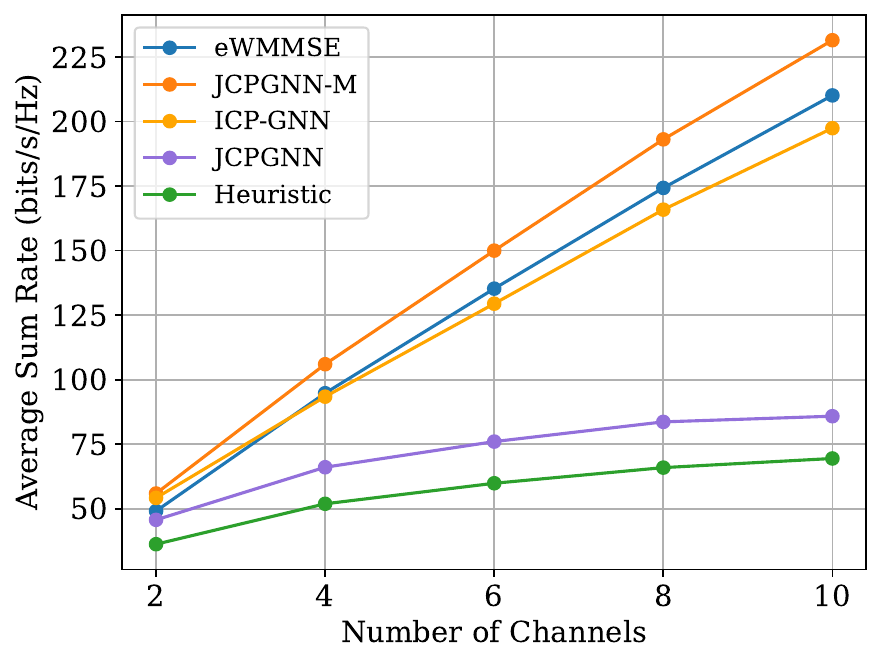}  
    \caption{$D=30$}
\end{subfigure}
\caption{Average sum rate comparison for different number of channels with $D=\{10,20,30\}$ scenarios.}
\label{C5:fig:srm_all}
\end{figure*}
\begin{subequations}\label{eq:aggreated}
\begin{align}
\boldsymbol{m}_{i,j}^{m (s)}&=\phi_1^{(s)}\left\{\left[\boldsymbol{x}_{i}^{m (s-1)},\mathbf{V}_{i}^m, \mathbf{E}_{i, j}^m\right]: j \in \mathcal{N}(v_i^m)\right\},\\
\boldsymbol{n}_{i}^{m (s)} &=\phi_2^{(s)}\left\{\left[\boldsymbol{m}_{i,j}^{m (s)}\right]\right\},
\end{align}
\end{subequations}
where $\boldsymbol{x}_{i}^{m}$ represents the optimization variable for vertex \(v_i^m\). The message computation process consists of two key steps:
First, a nonlinear transformation \(\phi_1^{(s)}(\cdot)\) is applied to generate the message \(\boldsymbol{m}_{i,j}^{m (s)}\). 
Then, an aggregation function \(\phi_2^{(s)}(\cdot)\) is used to consolidate the messages \(\boldsymbol{m}_{i,j}^{m (s)}\) from all neighbours, producing the aggregated message \(\boldsymbol{n}_{i}^{m (s)}\). We initialize $\boldsymbol{x}_{i}^{m}$ as \(\boldsymbol{x}_{i}^{m(0)} = 0\).

\subsubsection{Aggregation Layer}

After obtaining the aggregated message \(\boldsymbol{n}_{i}^{m(s)}\) from the message computation layer, the target vertex \(v_i^m\) updates its resource allocation. The update rule for aggregation in the \(s\)-th layer at vertex \(v_i^m\) is defined as:
\begin{equation}
\hat{p}_{i}^{m (s)}=\alpha^{(s)}\left(\boldsymbol{x}_{i}^{m (s-1)}, \boldsymbol{n}_{i}^{m(s)}\right),
\label{eq:gnnaggregate}
\end{equation}
where $\alpha^{(s)}(\cdot)$ is the update functions, typically implemented using MLPs. 
To ensure the power allocation values are within a valid range, the output layer of \(\alpha^{(s)}(\cdot)\) employs a Sigmoid activation function. 

\subsubsection{Post Processing Layer} 
The locally optimal solution either resides at a stationary point of \(\hat{\mathcal{L}}\) or on the boundary of the feasible region. Consequently, the following optimality conditions must be satisfied:
$\nabla_{\omega_p} \hat{\mathcal{L}} =0$, $\nabla_{\lambda_{i}}\hat{\mathcal{L}}=0$ (or $\lambda_i =0$). Since the maximum power constraint is a strict constraint, we aim to ensure that the solution satisfies \(\nabla_{\lambda_{i}} \hat{\mathcal{L}} \leq 0\). 
Therefore, we 
normalize the power allocations \(\hat{p}_{i}^{m}\) such that the total power equals \(P_{\max}\). Otherwise, no normalization is applied.

\section{Performance Evaluation}\label{C5:sec:performance_srm}
\subsection{Simulation Setup}
In this section, we conduct extensive simulations on the problem to evaluate the performance of the proposed JCPGNN-M framework. We adopt a system configuration that incorporates both large-scale fading and Rayleigh fading. In this scenario, we consider a system comprising $D$ transceiver pairs situated within a $100 \times 100$ $\text{m}^2$ area. The placement of transmitters is randomized within this area, and each receiver is randomly positioned around its transmitter. Their distance ranges from 2~m to 10~m. The hidden sizes of message computation $\phi_1(\cdot)$ and update functions $\alpha(\cdot)$ are $\{4,16,32\}$ and $\{33,16,8,1\}$, respectively. 
Here, we train our networks under 10000 training samples and test the performance with 1000 testing samples. Since the proposed JCPGNN-M algorithm is designed to solve Problem \eqref{eq:lagarian_gnn}, its primary objective is to minimize the Lagrangian function. The loss function for JCPGNN-M is Equation \eqref{C5:eq:LagrangeP}.

To assess the effectiveness of our proposed guideline, We consider three baselines for performance comparison as listed below,

\begin{itemize}
    \item ICP-GNN: Independent Channel Power Allocation GNN (ICP-GNN) is designed to independently allocate power to each channel \cite{shen2022graph}. Each user $i$ on channel $m$ can use a maximum power of $\frac{P_{\max}}{M}$. This ensures that power is evenly distributed across channels.
    \item JCPGNN: The GNN-based joint channel and power allocation algorithm proposed in \cite{chen2024gnn} is designed for a single-channel scenario, where each user is restricted to accessing only one channel at a time.
    \item Heuristic: We allocate the maximum transmit power $ P_{\max} $ to the channels where user pairs achieve the highest channel gain. For each user pair $ i $, we first identify the channel with the maximum channel gain, denoted as $ |h_{i,i}|_{\max} $, across all available channels. Then, we set $ p_i^m = P_{\max} $ if the channel gain $ |h_{i,i}^m| $ equals $ |h_{i,i}|_{\max} $. Otherwise, $ p_i^m $ is set to 0. 
    
\end{itemize}

\subsection{Performance Comparison}

We examine the sum rate maximization problem in Equation~\eqref{C5:eq:jointchannel}. The performance of different algorithms under the same simulation setup, evaluated across varying numbers of transceiver channels, is presented in Figure~\ref{C5:fig:srm_all}. 
We fix the number of pairs to $D=\{10,20,30\}$, and vary the number of channels to $M=\{2,4,6,8,10\}$. We observe that our proposed JCPGNN-M outperforms the state-of-the-art eWMMSE algorithm in all scenarios. When we fix the number of pairs, the performance gap increases with the number of channels. This is because JCPGNN-M can exploit the structure of wireless networks and allocate power more effectively with large number of channels. ICP-GNN performances are worse than eWMMSE with a larger number of channels. This is because the maximum power for each user in each channel is $\frac{P_{\max}}{M}$, which limits the potential of exploring the optimal solution. JCPGNN and the Heuristic algorithm have worse performance as they only allow one user to use one channel at a time, which limits the potential of fully using the channel resource.

\subsection{Generalization Capacity}
Apart from achieving a high sum rate performance, being able to generalize to larger scale problems is also important. In this section, we investigate the generalization capability of JCPGNN-M in the sum rate maximization problem.
\subsubsection{Generalization to Varied Network Densities}
We first train JCPGNN-M on a small network, e.g., $D=10$. Then, we test the trained networks with varied transceiver pair quantities while keeping the region dimension and number of channels consistent. The generalization performance is shown in Table~\ref{C5:Tab:generalize_users}. The performance has been normalized by the sum rate that the algorithm can achieve under the same set-up. Notably, our proposed JCPGNN-M achieves an impressive $99\%$ of the performance in most cases. 
\begin{table}[ht]
\centering
\caption{ Average generalization performance to different numbers of transceiver pairs, normalized by the performance achieved by JCPGNN-M in each setup.}
\begin{tabular}{|c|c|c|c|c|}
\hline \multicolumn{1}{|c|}{ System Scales } & $D=15$ & $D=20$ & $D=25$ & $D=30$ \\
\hline{$M=2$}  & $99.3\%$ & $99.1\%$ & $98.8\%$ & $98.7\%$  \\
\hline{$M=4$}   & $99.7\%$ & $99.4\%$ & $99.1\%$& $98.5\%$  \\
\hline{$M=6$}  & $99.9\%$ & $99.4\%$ & $99.7\%$  & $99.3\%$  \\
\hline{$M=8$}   & $99.8\%$ & $99.2\%$ & $98.9\%$& $98.4\%$ \\
\hline{$M=10$}   & $99.9\%$ & $99.3\%$ & $99.4\%$  & $98.6\%$ \\
\hline
\end{tabular}
\label{C5:Tab:generalize_users}
\end{table}
\begin{table}[ht]
\centering
\caption{Average generalization performance to different numbers of channels, normalized by the performance achieved by JCPGNN-M in each setup. }
\begin{tabular}{|c|c|c|c|c|c|}
\hline {System Scales} & $D=10$ & $D=15$ & $D=20$ & $D=25$& $D=30$ \\
\hline $M=4$  & $99.4\%$ & $99.6\%$ & $99.1\%$  & $99.6\%$ & $99.0\%$\\
\hline $M=6$  & $99.1\%$ & $99.5\%$ & $99.4\%$  & $99.9\%$ & $99.8\%$\\
\hline $M=8$  & $98.9\%$ & $99.4\%$ & $99.5\%$  & $99.1\%$ & $99.1\%$ \\
\hline $M=10$   & $98.3\%$ & $99.1\%$ & $98.9\%$  & $98.9\%$ & $98.9\%$\\
\hline
\end{tabular}
\label{C5:Tab:generalise_channels}
\end{table}
\subsubsection{Generalization to Varied Channel Number}
Similarly, we first train JCPGNN-M using the two-channel case and then evaluate its performance on scenarios with varying numbers of channels. 
The results, presented in Table~\ref{C5:Tab:generalise_channels}, demonstrate that our proposed JCPGNN-M maintains strong generalization performance. Notably, it achieves $98\%$ of the best performance even when tested on configurations where the number of channels is increased by five times, highlighting its adaptability to different channel conditions.

\subsection{Time Complexity}
\begin{figure}
\centerline{\includegraphics[width =.85\linewidth]{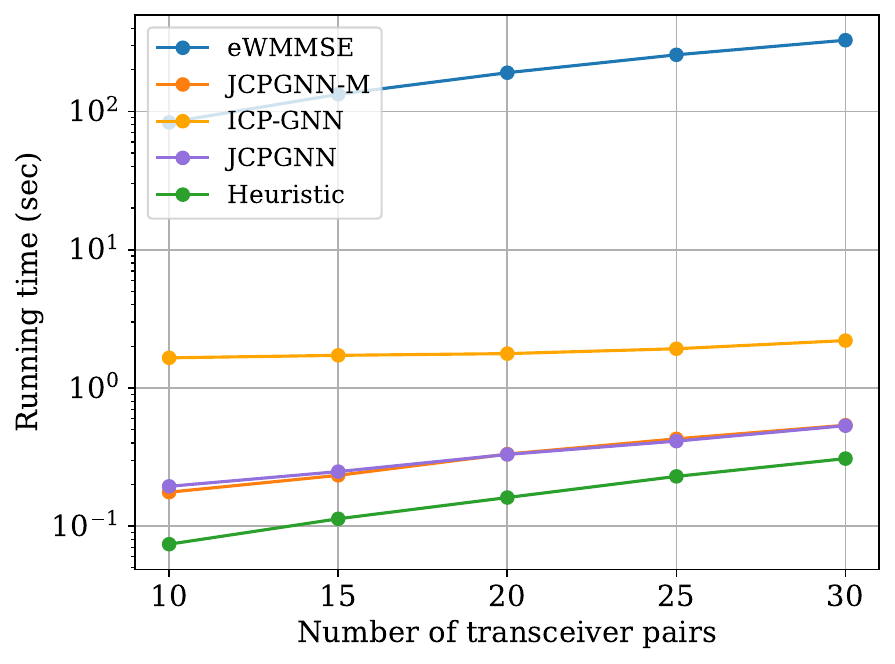}}
\caption{Average running time of different algorithms when $M=10$ (the figure is presented on a logarithmic scale).}
\label{C5:fig:srm_all_time}
\end{figure}

The average running time of the algorithms, under the same experimental setup as in Figure~\ref{C5:fig:srm_all}, is presented in Figure~\ref{C5:fig:srm_all_time}. The heuristic algorithm has the lowest computational complexity due to its simplicity and minimal processing requirements. Our proposed JCPGNN-M demonstrates significantly lower complexity compared to ICP-GNN and the state-of-the-art eWMMSE algorithm. ICP-GNN has a higher computational cost as it processes each channel individually, leading to an increasing performance gap to JCPGNN-M as the total number of channels grows. 
Notably, JCPGNN-M achieves better performance than eWMMSE while requiring only $1\%$ of its running time, demonstrating its superior efficiency. 

\section{Conclusion}\label{C5:sec:conclusion}
This paper presents a joint channel and power allocation framework for multi-channel wireless networks, addressing the limitations of conventional optimization techniques. To efficiently solve the problem, we propose a GNN-based algorithm, JCPGNN-M, which allows users to access multiple channels simultaneously. This approach enhances spectral efficiency and overall network performance.

\bibliographystyle{IEEEtran}
\bibliography{main}

\begin{thebibliography}{10}
\providecommand{\url}[1]{#1}
\csname url@samestyle\endcsname
\providecommand{\newblock}{\relax}
\providecommand{\bibinfo}[2]{#2}
\providecommand{\BIBentrySTDinterwordspacing}{\spaceskip=0pt\relax}
\providecommand{\BIBentryALTinterwordstretchfactor}{4}
\providecommand{\BIBentryALTinterwordspacing}{\spaceskip=\fontdimen2\font plus
\BIBentryALTinterwordstretchfactor\fontdimen3\font minus \fontdimen4\font\relax}
\providecommand{\BIBforeignlanguage}[2]{{%
\expandafter\ifx\csname l@#1\endcsname\relax
\typeout{** WARNING: IEEEtran.bst: No hyphenation pattern has been}%
\typeout{** loaded for the language `#1'. Using the pattern for}%
\typeout{** the default language instead.}%
\else
\language=\csname l@#1\endcsname
\fi
#2}}
\providecommand{\BIBdecl}{\relax}
\BIBdecl

\bibitem{shi2011iteratively}
Q.~Shi, M.~Razaviyayn, Z.-Q. Luo, and C.~He, ``An iteratively weighted {MMSE} approach to distributed sum-utility maximization for a {MIMO} interfering broadcast channel,'' \emph{IEEE Transactions on Signal Processing}, vol.~59, no.~9, pp. 4331--4340, 2011.

\bibitem{elnourani2018underlay}
M.~Elnourani, M.~Hamid, D.~Romero, and B.~Beferull-Lozano, ``Underlay device-to-device communications on multiple channels,'' in \emph{2018 IEEE International Conference on Acoustics, Speech and Signal Processing (ICASSP)}.\hskip 1em plus 0.5em minus 0.4em\relax IEEE, 2018, pp. 3684--3688.

\bibitem{hajiaghajani2016joint}
F.~Hajiaghajani, R.~Davoudi, and M.~Rasti, ``A joint channel and power allocation scheme for device-to-device communications underlaying uplink cellular networks,'' in \emph{2016 IEEE Conference on Computer Communications Workshops (INFOCOM WKSHPS)}.\hskip 1em plus 0.5em minus 0.4em\relax IEEE, 2016, pp. 768--773.

\bibitem{mach2019resource}
P.~Mach, Z.~Becvar, and M.~Najla, ``Resource allocation for {D2D} communication with multiple {D2D} pairs reusing multiple channels,'' \emph{IEEE Wireless Communications Letters}, vol.~8, no.~4, pp. 1008--1011, 2019.

\bibitem{peng2024learning}
Y.~Peng, T.~Liu, and C.~Yang, ``Learning power allocation for cell-free massive mimo system with graph neural networks,'' in \emph{GLOBECOM 2024-2024 IEEE Global Communications Conference}.\hskip 1em plus 0.5em minus 0.4em\relax IEEE, 2024, pp. 2653--2658.

\bibitem{chen2023graph}
L.~Chen, J.~Zhu, and J.~Evans, ``Graph neural networks for power allocation in wireless networks with full duplex nodes,'' in \emph{2023 IEEE International Conference on Communications Workshops (ICC Workshops)}, 2023, pp. 277--282.

\bibitem{li2024hetero}
B.~Li, L.-L. Yang, R.~G. Maunder, S.~Sun, and P.~Xiao, ``Heterogeneous graph neural network for power allocation in multicarrier-division duplex cell-free massive mimo systems,'' \emph{IEEE Transactions on Wireless Communications}, vol.~23, no.~2, pp. 962--977, 2024.

\bibitem{chen2021gnn}
T.~Chen, X.~Zhang, M.~You, G.~Zheng, and S.~Lambotharan, ``A {GNN}-based supervised learning framework for resource allocation in wireless iot networks,'' \emph{IEEE Internet of Things Journal}, vol.~9, no.~3, pp. 1712--1724, 2021.

\bibitem{chen2024gnn}
L.~Chen, J.~Zhu, and J.~Evans, ``{GNN}-based joint channel and power allocation in heterogeneous wireless networks,'' in \emph{2024 IEEE International Conference on Communications Workshops (ICC Workshops)}.\hskip 1em plus 0.5em minus 0.4em\relax IEEE, 2024, pp. 233--238.

\bibitem{marwani2024graph}
M.~Marwani and G.~Kaddoum, ``Graph neural networks approach for joint wireless power control and spectrum allocation,'' \emph{IEEE Transactions on Machine Learning in Communications and Networking}, 2024.

\bibitem{nakashima2020deep}
K.~Nakashima, S.~Kamiya, K.~Ohtsu, K.~Yamamoto, T.~Nishio, and M.~Morikura, ``Deep reinforcement learning-based channel allocation for wireless {LAN}s with graph convolutional networks,'' \emph{IEEE Access}, vol.~8, pp. 31\,823--31\,834, 2020.

\bibitem{sun2018learning}
H.~Sun, X.~Chen, Q.~Shi, M.~Hong, X.~Fu, and N.~D. Sidiropoulos, ``Learning to optimize: Training deep neural networks for interference management,'' \emph{IEEE Transactions on Signal Processing}, vol.~66, no.~20, pp. 5438--5453, 2018.

\bibitem{keriven2019universal}
N.~Keriven and G.~Peyr{\'e}, ``Universal invariant and equivariant graph neural networks,'' \emph{Advances in Neural Information Processing Systems}, vol.~32, 2019.

\bibitem{sun2023unsupervised}
C.~Sun, C.~She, and C.~Yang, ``Unsupervised deep learning for optimizing wireless systems with instantaneous and statistic constraints,'' \emph{Ultra-Reliable and Low-Latency Communications (URLLC) Theory and Practice: Advances in 5G and Beyond}, pp. 85--117, 2023.

\bibitem{eisen2019learning}
M.~Eisen, C.~Zhang, L.~F. Chamon, D.~D. Lee, and A.~Ribeiro, ``Learning optimal resource allocations in wireless systems,'' \emph{IEEE Transactions on Signal Processing}, vol.~67, no.~10, pp. 2775--2790, 2019.

\bibitem{shen2022graph}
Y.~Shen, J.~Zhang, S.~Song, and K.~B. Letaief, ``Graph neural networks for wireless communications: From theory to practice,'' \emph{IEEE Transactions on Wireless Communications}, vol.~22, no.~5, pp. 3554--3569, 2022.

\end{thebibliography}

\vspace{12pt}

\end{document}